\documentclass{article}
\pdfoutput=1

\usepackage{arxiv}

\usepackage[utf8]{inputenc} 
\usepackage[T1]{fontenc}    
\usepackage{hyperref}       
\usepackage{url}            
\usepackage{booktabs}       
\usepackage{amsfonts}       
\usepackage{nicefrac}       
\usepackage{microtype}      
\usepackage{lipsum}		
\usepackage{graphicx}
\usepackage{natbib}
\usepackage{doi}

\usepackage{colortbl}
\usepackage{tabulary}
\usepackage{etoolbox}

\title{A Rapid Review of Responsible AI frameworks: How to guide the development of ethical AI}

\author{ \href{https://orcid.org/0000-0002-0163-6786}{\includegraphics[scale=0.06]{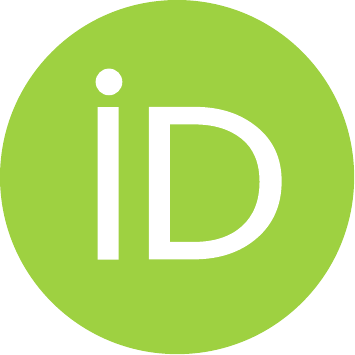}\hspace{1mm}Vita Santa Barletta} \\
	Department of Computer Science\\
	University of Bari "A. Moro"\\
	Bari, Italy \\
	\texttt{vita.barletta@uniba.it} \\
	\And
	\href{https://orcid.org/0000-0001-5719-7447}{\includegraphics[scale=0.06]{orcid.pdf}\hspace{1mm}Danilo Caivano} \\
	Department of Computer Science\\
	University of Bari "A. Moro"\\
	Bari, Italy \\
	\texttt{danilo.caivano@uniba.it} \\
	\And
	\href{https://orcid.org/0000-0003-3589-6970}{\includegraphics[scale=0.06]{orcid.pdf}\hspace{1mm}Domenico Gigante} \\
	Department of Computer Science\\
	University of Bari "A. Moro"\\
	Bari, Italy \\
	\texttt{domenico.gigante1@uniba.it} \\
	\And
	\href{https://orcid.org/0000-0002-3537-7663}{\includegraphics[scale=0.06]{orcid.pdf}\hspace{1mm}Azzurra Ragone} \\
	Department of Computer Science\\
	University of Bari "A. Moro"\\
	Bari, Italy \\
	\texttt{azzurra.ragone@uniba.it} \\
}

\renewcommand{\shorttitle}{A Rapid Review of Responsible AI frameworks}

\hypersetup{
pdftitle={A Rapid Review of Responsible AI frameworks: How to guide the development of ethical AI},
pdfsubject={Responsible AI},
pdfauthor={Vita Santa Barletta, Danilo Caivano, Domenico Gigante, Azzurra Ragone},
pdfkeywords={artificial intelligence, software engineering, responsible ai, development frameworks, literature review},
}

\begin{document}
\maketitle

\begin{abstract}
	In the last years, the raise of Artificial Intelligence (AI), and its pervasiveness in our lives, has sparked a flourishing debate about the ethical principles that should lead its implementation and use in society. Driven by these concerns, we conduct a rapid review of several frameworks providing principles, guidelines, and/or tools to help practitioners in the development and deployment of Responsible AI (RAI) applications. We map each framework w.r.t. the different Software Development Life Cycle (SDLC) phases discovering that most of these frameworks fall just in the \textit{Requirements Elicitation} phase, leaving the other phases uncovered. Very few of these frameworks offer supporting tools for practitioners, and they are mainly provided by private companies. Our results reveal that there is not a "catching-all" framework supporting both technical and non-technical stakeholders in the implementation of real-world projects. Our findings highlight the lack of a comprehensive framework encompassing all RAI principles and all (SDLC) phases that could be navigated by users with different skill sets and with different goals.
\end{abstract}

\keywords{artificial intelligence \and software engineering \and responsible ai \and development frameworks \and literature review}

\section{Introduction}
\label{sec:intro}

Artificial Intelligence (AI) is a revolution that is reshaping science and society as a whole \cite{Harari2017}. While AI-related technologies 
are changing how data is processed and analyzed \cite{10.1126/science.aaa8415}, autonomous and semi-autonomous decision systems are being used more frequently in several industries, such as healthcare, automotive, banking, and manufacturing, just to cite a few \cite{CornacchiaNR21}. Given AI revolutionary potential and wide-ranging social influence, there has been a lot of discussion regarding the values and principles that should lead its development and application \cite{Vayena2018}\cite{Awad2018}. Recent scientific research and media attention have been focused on concerns that AI may endanger the jobs of human workers \cite{naturenews2017}, be abused by malicious actors \cite{Brundage2018}, avoid responsibility, or accidentally spread bias and as so, erode fairness \cite{Zou2018AICB}.

In this context, the concept of Responsible Artificial Intelligence (RAI) was defined: "\textit{intelligent algorithms that prioritize the needs of all stakeholders as the highest priority, especially the minoritized and disadvantaged users, in order to make trustworthy decisions. These obligations include protecting and informing users, preventing and mitigating negative impacts, and maximizing the long-term beneficial impact. (Socially) Responsible AI Algorithms constantly receive feedback from users to continually accomplish the expected social values}" \cite{cheng2021socially}. Several public and private organizations have responded to these societal fears by developing different kinds of resources: ethical requirements, principles, guidelines, best practices, tools, and frameworks.
In this work, we conducted a Rapid Review (RR) of various frameworks addressing the principles of RAI proposed in the literature by different types of institutions. In particular, we investigated frameworks giving practical guidance and support to all types of stakeholders involved in the implementation and validation of AI applications, also with reference to the Software Development Life Cycle (SDLC). 

In this work we investigated the state of the practice on how comprehensive and complete RAI frameworks are in terms of principles addressed and SDLC phases covered, as well as if there are tools complementing the theoretical framework helping practitioners in the different phases of the development lifecycle. Following to our Rapid Review, our main finding is that there is a concerning shortage of tools supporting the design, implementation, and auditing of the RAI principles both by technical and non-technical stakeholders. Further research should focus on the development of a comprehensive and simple-to-use framework for RAI whose knowledge can be navigated and exploited by AI practitioners to speed up the adoption of RAI practices in real-world projects.

The remainder of the paper is organized as follows: Section \ref{sec:background} defines some background definitions useful for this study; Section \ref{sec:Study Design} describes the protocol we adopted to conduct the rapid review, including research questions and the research strategy adopted; Section \ref{sec:results} provides the results of our rapid review and answers the research questions; Section \ref{sec:discussion} discusses the results highlighting the most important findings; Section \ref{sec:threats} addresses threats to validity and finally the conclusion and future work are drawn in Section \ref{sec:conclusions}.
\section{Background}
\label{sec:background}

We provide some preliminary definitions to better understand the concepts that guided this work.

\subsection{Responsible AI Principles}
\label{subsec:ai_principles}

National and international organizations have created ad-hoc expert groups on AI to address the risks connected with the development of AI, frequently with the task of generating policy documents. 
These organizations include, among others, the High-Level Expert Group on Artificial Intelligence established by the European Commission\footnote{\url{https://digital-strategy.ec.europa.eu/en/policies/expert-group-ai}}, the UNESCO Ad Hoc Expert Group (AHEG) for the Recommendation on the Ethics of Artificial Intelligence\footnote{\url{https://www.unesco.org/en/artificial-intelligence/recommendation-ethics}}, the Advisory Council on the Ethical Use of Artificial Intelligence and Data in Singapore\footnote{\url{https://www.cms-holbornasia.law/en/sgh/publication/singapore-to-form-advisory-council-for-ethical-use-of-ai}}, the NASA Artificial Intelligence Group\footnote{\url{https://ai.jpl.nasa.gov/}} and the UK AI Council\footnote{\url{https://www.gov.uk/government/groups/ai-council}}, just to cite a few.

These committees have been appointed to produce reports and guidelines about RAI. Similar initiatives are being made in the commercial sector, particularly by businesses that depend on AI. Businesses like Sony\footnote{\url{https://www.sony.com/en/SonyInfo/sony_ai/responsible_ai.html}} and Meta\footnote{\url{https://ai.facebook.com/blog/facebooks-five-pillars-of-responsible-ai/}} made their AI policies and principles available to the public. 
At the same time, professional organizations and no-profit groups like UNI Global Union\footnote{\url{http://www.thefutureworldofwork.org/media/35420/uni_ethical_ai.pdf}} and the Internet Society\footnote{\url{https://www.internetsociety.org/resources/doc/2017/artificial-intelligence-and-machine-learning-policy-paper/}} have all released statements and recommendations.

The significant efforts of such an ample group of stakeholders to develop RAI principles and policies not only show the need for ethical guidance but also point out their keen interest in reshaping AI ethics to suit their individual priorities \cite{Greene2019BetterNC}. Notably, the private sector's participation in the field of AI ethics has been questioned since it may be using high-level soft policy as a portmanteau to either make a social issue technical \cite{Greene2019BetterNC} or avoid regulation altogether \cite{bayamlioglu_being_2018, Jobin2019}.

However, many research works highlighted how these proposals often diverged, giving different definitions, resulting in the problem known as \textit{principle proliferation} \cite{Floridi2019AUF}. Consequently, several in-depth investigations have been conducted, such as the one by Jobin et al. \cite{Jobin2019}, who found a
global convergence around five ethical principles: \textit{transparency}, \textit{justice and fairness}, \textit{non-maleficence}, \textit{responsibility}, and \textit{privacy}. 
Jobin et al. \cite{Jobin2019} stated that no one of these ethical principles is present in all the documents they reviewed; however, these five principles are mentioned in more than half of all the sources reviewed. Moreover, further in-depth thematic analysis revealed notable semantic and conceptual divergences in interpreting these principles and in the particular recommendations or areas of concern drawn from each of them.

\subsection{Chosen AI principles definitions}
\label{subsec:chosen_definition}
As highlighted in Section \ref{subsec:ai_principles}, there are a lot of uncertainties and nuances around the definition of the principles that mainly characterize Responsible AI, as well as, about the definition of RAI itself. Indeed, sometimes it is referred to as Trustworthy or Ethical AI. 
In our work, we address the problem of \textit{principle proliferation} deciding to focus on a specific subset of those characterizing RAI, in particular, the four principles identified by Jobin et al. \cite{Jobin2019} with the exclusion of \textit{responsibility} as this concept is rarely defined in a clear manner.

Moreover, to give an authoritative and clear definition for each principle, we decided to use the ones provided by the High-Level Expert Group on Artificial Intelligence established by the European Commission\footnote{\url{https://digital-strategy.ec.europa.eu/en/policies/expert-group-ai}} in their \textit{Ethics guidelines for trustworthy AI} \cite{AIHLEG_ethics_guidelines}.

In the following we report the selected definitions for each principle. We matched the principles in \cite{AIHLEG_ethics_guidelines} with the ones that emerged in \cite{Jobin2019}, and when the naming convention is not exactly the same we highlighted that in parenthesis.
We also mapped principles to system requirements.



\begin{figure}[h]
\centering
\includegraphics[scale=1]{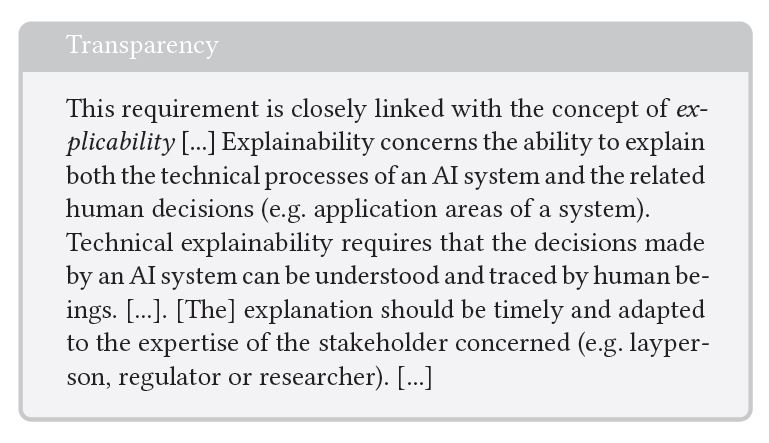}
\label{fig:transparency_box}
\end{figure}

This principle includes and can be directly linked with system requirements such as \textit{traceability} and \textit{explainability}.


\begin{figure}[h]
\centering
\includegraphics[scale=1]{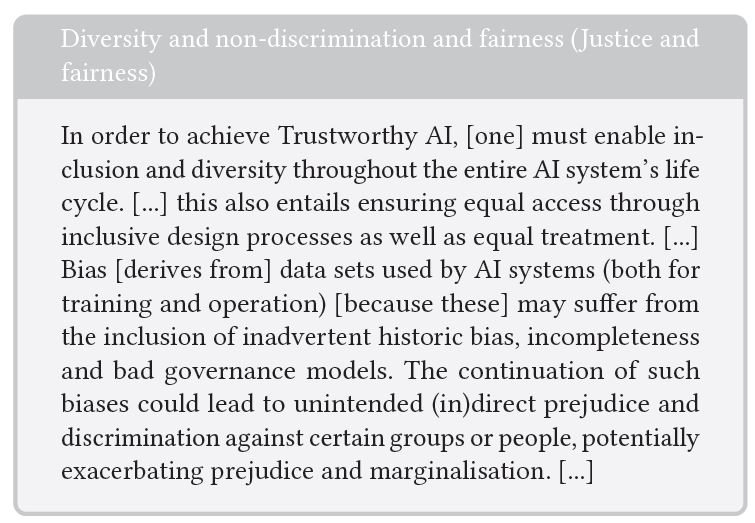}
\label{fig:fairness_box}
\end{figure}

This principle can be directly mapped with system requirements such as \textit{avoidance of unfair bias}, \textit{accessibility and universal design} and \textit{include stakeholder participation}.


\begin{figure}[h]
\centering
\includegraphics[scale=1]{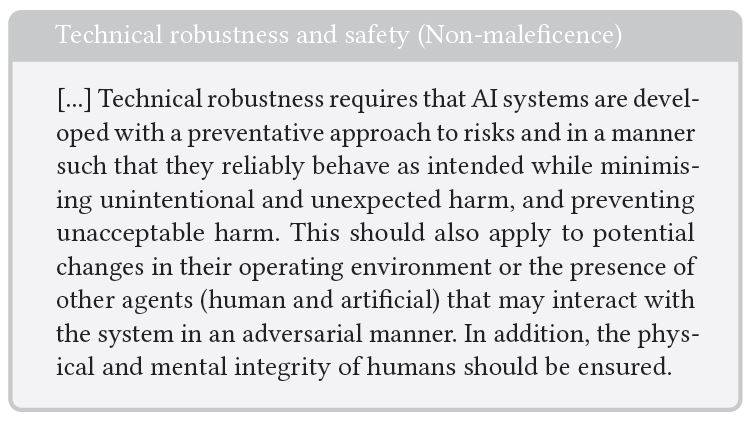}
\label{fig:robustness_box}
\end{figure}

This principle can be directly mapped with system requirements such as \textit{resilience to attack and security}, \textit{fallback plan} and \textit{general safety, accuracy, reliability}, and \textit{reproducibility}.


\begin{figure}[h]
\centering
\includegraphics[scale=1]{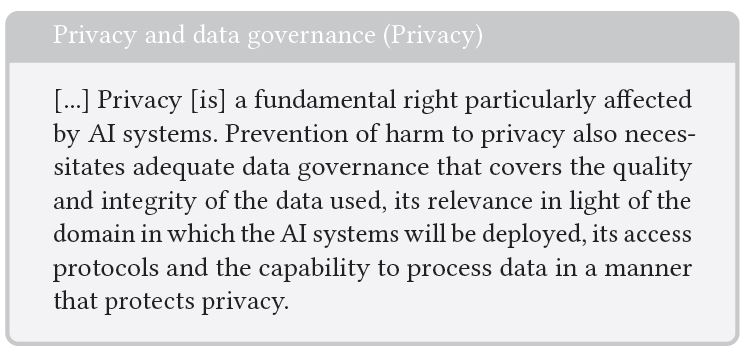}
\label{fig:privacy_box}
\end{figure}

This principle can be directly mapped with system requirements such as \textit{including respect for privacy}, \textit{quality and integrity of data} and \textit{quality and integrity of access to data}.

\subsection{Frameworks}
\label{subsec:frameworks}
In this work, we focus on frameworks that implement the above-mentioned RAI ethical principles.

The concept of \textit{framework} is far well-known in the Software Engineering (SE) field. Already in 1997, Johnson et al. \cite{Johnson199739} referred to frameworks as "\textit{an object-oriented reuse technique}" or "\textit{the skeleton of an application that can be customized by an application developer}". These are not conflicting definitions; the first describes the structure of a framework while the second describes its purpose. 

Shifting the focus from SE to a more general context, frameworks are a form of design reuse. 
Frameworks can be considered a collection of suggestions, guidelines and tools to be followed in order to create a product compliant with a defined standard.
\section{Study Design}
\label{sec:Study Design}

Rapid Reviews (RRs) have emerged as a streamlined approach for synthesizing evidence quickly, initially intending to help decision-makers in health care to respond promptly to urgent and emerging needs \cite{konnyu2012}. Rapid reviews simplify systematic review methods by focusing on the literature search while still aiming to produce valid conclusions \cite{wattcameron2008}.

To perform this rapid review, we followed the protocol proposed in \cite{10.1145/3210459.3210462}, and we complemented the Rapid Review process with the strategies presented in \cite{Kitchenham2007} for performing systematic literature reviews. The following subsections describe in detail the study design and its execution.

\subsection{Planning the review}
\label{subsec:planning_the_review}
The rapid literature review presented in this work was carried out through the following steps:
\begin{enumerate}
    \item \textbf{Goal and Research questions}: the goal and the correlated research questions were identified to guide the literature review;
    \item \textbf{Search strategy}: defining the strategy to collect previous works published in the literature, including research databases and query strings;
    \item \textbf{Eligibility criteria definition}: the criteria used to filter the collected studies have been defined;
    \item \textbf{Data extraction}: defining how relevant data were extracted to help answer the research questions;
    \item \textbf{Data synthesis}: defining how to organize extracted relevant data to answer the research questions.
\end{enumerate}

\begin{figure*}[ht]
\centering
\includegraphics[scale=0.6]{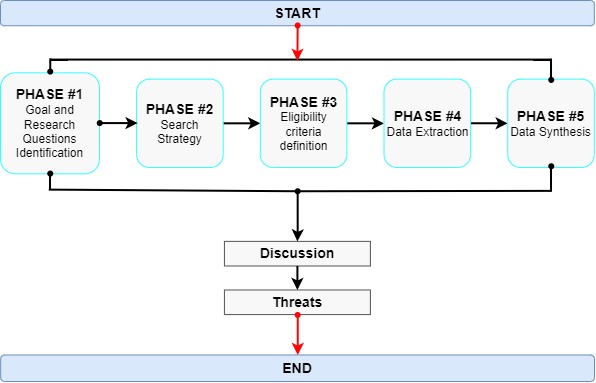}
\caption{Research protocol used in this rapid review.}
\label{fig:research_protocol}
\end{figure*}

Fig. \ref{fig:research_protocol} summarizes the review protocol.

\subsection{Goal and Research Questions}
\label{subsec:goal_and_rq}

The goal of our study is to gather, organize, and analyze the RAI frameworks proposed in the literature by public and private institutions to investigate, first, how much comprehensive they are in terms of principles addressed and SDLC phases covered. Then, we focus specifically on frameworks offering tools that can be used in the implementation and auditing phase of the decision systems offering practical guidance to practitioners.

Based on this goal, we defined the following research questions:
\begin{itemize}
    \item \textbf{RQ1}: What are the Responsible AI frameworks proposed in the literature?
    \item \textbf{RQ2}: How much do these frameworks address the various RAI principles?
    \item \textbf{RQ3}: Do these frameworks provide recommendations for each phase of the Software Development Life Cycle (SDLC)?
    \item \textbf{RQ4}: Is there a supporting tool for each proposed framework?
\end{itemize}

We recall that we decided to focus our research on the most cited RAI principles \cite{Jobin2019}:

\textbf{Transparency}, \textbf{Diversity \& Non-discrimination \& Fairness}, \textbf{Technical Robustness \& Safety},  and \textbf{Privacy \& Data Governance}.

\subsection{Search strategy}
\label{subsec:search_strategy}

As recommended in \cite{10.1145/3210459.3210462}, to abbreviate the search for primary studies and conduct the rapid review within the available time, we used only \textbf{Scopus}\footnote{\url{https://www.scopus.com/}} and \textbf{Google Scholar}\footnote{\url{https://scholar.google.com/}} as white literature search engines.

To improve the search string, we conducted pilot searches and we excluded those keywords that did not yield coherent search results (e.g., the \textit{knowledge base} keyword). After some trials, we found the search string presented in the following box that returned several relevant papers.


\begin{figure}[ht]
\centering
\includegraphics[scale=1]{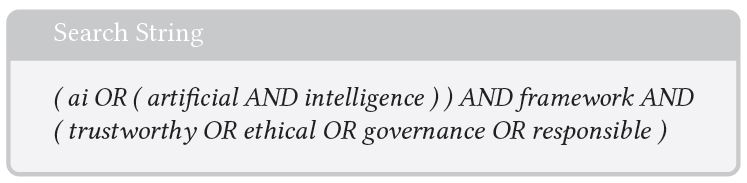}
\label{fig:search_string}
\end{figure}

The selection string was run on 15 November 2022 at 12:40 and initially produced 1875 document results on Scopus. Then, some filters were applied: we selected only "Computer Science", "Social Sciences", "Engineering" and "Mathematics" subject areas and selected only documents written in English or Italian; these filters reduced the total amount of results to 1489. After a few days, the same string has been run on Google Scholar - filtering the results by "Review articles" type - and produced 91200 results.

The classification process of the documents collected by Google Scholar stopped after 20 pages because, starting from the 21st, the documents did not include all the keywords contained in the search string and/or were not coherent with the research objective; a total of 200 documents collected by Google Scholar were analyzed.

Later, to further enhance the quality of our research, we moved to analyze grey-literature sources. First of all, \textbf{Algorithm Watch} AI Ethics Guidelines Global Inventory\footnote{\url{https://algorithmwatch.org/en/ai-ethics-guidelines-global-inventory/}} database was queried to identify the business companies which proposed RAI resources. Here, three types of resources were discarded:
\begin{itemize}
    \item The ones whose link was no more working and the same resource could be not found even searching on Google.com search engine;
	\item Unofficial documents, e.g. unofficial translations;
	\item Documents not written in English or Italian.
\end{itemize}
In this stage, for each resource found on Algorithm Watch, an in-depth investigation has been conducted on the proponent entity's website to discover useful resources (and frameworks) not indexed on Algorithm Watch AI Ethics Guidelines Global Inventory.

Then, the \textbf{OECD database}\footnote{\url{https://oecd.ai/en/catalogue/tools}} was queried as well. Here, some filters were applied: 
\begin{itemize}
    \item \textit{Procedural} Category, which (according to the description) is the most similar to the concept of "framework";
	\item \textit{Published document} or \textit{implemented in multiple projects} as Tool Readiness;
	\item \textit{Fairness, Privacy \& data governance, Robustness \& digital security, Transparency \& explainability} as Object.
\end{itemize}
Finally, a keyword-based search on Google was performed in private-browsing mode, after logging out from personal accounts and erasing all web cookies and history \cite{Piasecki2017}. The search was performed using the same search string used on Scopus and Scholar. This last search initially produced 21,100,000 results but, after manually inspecting the results contained in the first 17 pages---for a total amount of 168 resources---, the search engine hid subsequent results since they were very similar to the previously inspected ones.
From the Google search results, we also discarded all the duplicates of the documents already retrieved from other data sources.

All the documents obtained with this search strategy were surveyed using a 3-stages information classification process. 
In the first stage, only the title and keywords of the collected articles were read. In the second stage, we analyzed the abstract of each article while in the third stage we read the complete article. All these stages were conducted separately and in blind-view way by two of the authors. In case of a disagreement, a third author manually verified and took the final decision.

\subsection{Eligibility criteria definition}
\label{subsec:selection_criteria}

The selection procedure was based on the following criteria:
\begin{enumerate}
    \item The resource must be in English or Italian;
    \item The resource must be in the context of Responsible AI frameworks;
    \item The resource must address at least one of the chosen principles (see Section \ref{subsec:chosen_definition});
    \item The resource must provide answers to at least one of the rapid review’s research questions.
\end{enumerate}
The full-text of every collected resource was validated against these criteria and, if everyone was satisfied, the resource was surveyed. The complete results list can be seen in the online appendix \cite{online_appendix}.

\subsection{Data Extraction}
\label{subsec:extraction_procedure}

In this step, we extracted all relevant data that could help answer any of the research questions. 
The extraction process was performed by two of the authors and conflicts were solved by a third author in a blind-view way. We used a worksheet to tabulate and organize data \cite{online_appendix}.
\section{Results}
\label{sec:results}
The RR was performed part-time from 15 November 2022 until 12 December 2022, conducting the procedure described in Section \ref{subsec:search_strategy}.

\begin{table}[h]
\centering
\caption{Amount of documents collected grouped by research phase.}
\label{table:research_results}
\begin{tabular}{|c|c|c|c|}
\hline
\textbf{\begin{tabular}[c]{@{}c@{}}\textbf{Data} \\ \textbf{Source}\end{tabular}} & \textbf{\begin{tabular}[c]{@{}c@{}}\textbf{Resources} \\ \textbf{retrieved}\end{tabular}} & \textbf{\begin{tabular}[c]{@{}c@{}}\textbf{Resources} \\ \textbf{analyzed}\end{tabular}} & \textbf{\begin{tabular}[c]{@{}c@{}}\textbf{Resource} \\ \textbf{selected}\end{tabular}} \\ \hline
Scopus                                                          & 1875                                  & 1489                            & 20                                                                     \\ \hline
Google Scholar                                                  & 91200                                 & 200                             & 0                                                                     \\ \hline
Algorithm Watch                                                 & 167                                   & 167                             & 80                                                                  \\ \hline
OECD DB                                                         & 356                                    & 70                              & 38                                                                    \\ \hline
Google Search                                                   & 21,100,00                             & 168                             & 10                                                                    \\ \hline
\end{tabular}
\end{table}

Table \ref{table:research_results} summarises the number of relevant results obtained in each sub-phase. The search in Google Scholar did not produce any useful results. 
Summing up the resources selected from the identified data sources in the last step, we ended up with 148 unique resources (without duplicates).

\subsection{Data synthesis}
\label{subsec:data_synthesis}
Data were synthesized and several statistics were computed to answer the research questions (see the online appendix available at \cite{online_appendix}). In Tables \ref{table:overall_results_by_principle} and \ref{table:overall_results_by_sdlc} we show an illustrative excerpt of the whole data collected.

\begin{table*}[t]
\caption{Excerpt of the whole data classified by RAI principles taken into consideration.}
\begin{center}
\begin{tabular}{c|l|c|c|c|c}
\hline
\textbf{Entity name} & \textbf{Tool Name} & \textbf{\begin{tabular}[c]{@{}c@{}}Diversity \\ Non-discrimination \\ \& Fairness\end{tabular}} & \textbf{\begin{tabular}[c]{@{}c@{}}Privacy \\ \& Data \\ Governance\end{tabular}} & \textbf{\begin{tabular}[c]{@{}c@{}}Technical \\ Robustness \\ \& Safety\end{tabular}} & \textbf{\begin{tabular}[c]{@{}c@{}}Transparency \end{tabular}} \\
\hline
\multicolumn{6}{c}{\cellcolor{yellow}\textbf{COMPANIES}}                                                                                                                                                                                                                                                                                                                                                                                                       \\
Meta                                                                   & \begin{tabular}[c]{@{}l@{}}Facebook’s five pillars \\ of Responsible AI\end{tabular}        & Yes               & Yes                                                                            & Yes                                                                                & Yes                                                                                \\
\multicolumn{6}{c}{\cellcolor{yellow}\textbf{UNIVERSITIES}}                                                                                                                                                                                                                                                                                                                                                                                                    \\
\begin{tabular}[c]{@{}l@{}}University of \\ Texas at Austin\end{tabular} & \begin{tabular}[c]{@{}l@{}}CERTIFAI\end{tabular} & Yes                & No                                                                             & Yes                                                                                 & Yes                                                                                 \\
\multicolumn{6}{c}{\cellcolor{yellow}\textbf{NO-PROFIT ORG / COMMUNITIES / GOVERNMENT ENTITIES}}                                                                                                                                                                                                                                                                                                                                                               \\
NIST                                                                    	& \begin{tabular}[c]{@{}l@{}}AI Risk \\ Management Framework\end{tabular}        & Yes               & Yes                                                                            & Yes                                                                                & Yes                                                                                \\
\end{tabular}
\label{table:overall_results_by_principle}
\end{center}
\end{table*}

\begin{table}[t]
\centering
\caption{Excerpt of the whole data classified by SDLC phase addressed.}
\begin{center}
\begin{tabular}{c|c|c|c|c|c}
\hline
\textbf{Entity name}                                                                            & \textbf{\begin{tabular}[c]{@{}c@{}}Requirements \\ Elicitation\end{tabular}} & \textbf{Design} & \textbf{Development} & \textbf{Testing} & \textbf{Deployment} \\
\hline
\multicolumn{6}{c}{\cellcolor{yellow}\textbf{COMPANIES}}                                                                                                                                                                                                   \\
Meta                                                                    & Yes                                                                & No   & No        & No    & No       \\
\multicolumn{6}{c}{\cellcolor{yellow}\textbf{UNIVERSITIES}}                                                                                                                                                                                                \\
\begin{tabular}[c]{@{}l@{}}University of \\ Texas at Austin \\ (CERTIFAI)\end{tabular} & No                                                                           & Yes              & Yes                   & Yes               & No                  \\
\multicolumn{6}{c}{\cellcolor{yellow}\textbf{NO-PROFIT ORG / COMMUNITIES / PUBLIC ENTITIES}}                                                                                                                                                               \\
NIST                                                                        & Yes                                                                          & Yes             & No                   & No               & No                 
\end{tabular}
\label{table:overall_results_by_sdlc}
\end{center}
\end{table}

\textbf{RQ1.} \emph{What are the Responsible AI frameworks proposed in the literature?}

All the retrieved frameworks have been classified w.r.t. the \textbf{type of proposing institution} using three categories: COMPANIES, UNIVERSITIES, and NO-PROFIT ORG / COMMUNITIES / PUBLIC ENTITIES (NPG/COMM/PE).
 Moreover, if an entity proposed two (or more) frameworks, it was counted only once, as we are interested in the distribution of the proposals by entity type.

\begin{figure}[ht]
\centering
\includegraphics[scale=0.7]{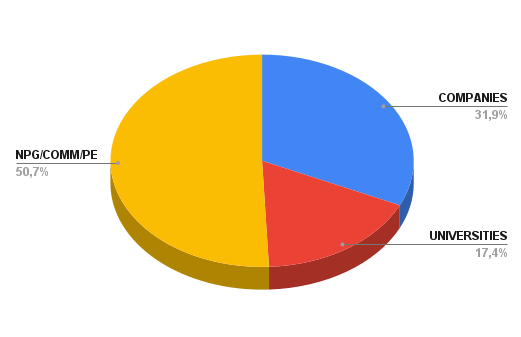}
\caption{Distribution by type of proposing institution.}
\label{fig:research_results_type_proposing_institution_distribution}
\end{figure}

As can be seen in Fig. \ref{fig:research_results_type_proposing_institution_distribution}, what immediately emerges is that most of the filtered frameworks are proposed by NPG/COMM/PE (50.7\%, 70/138), followed by lucrative COMPANIES (31.9\%, 44/138) and then UNIVERSITIES (17.4\%, 24/138).

\begin{figure}[ht]
\centering
\includegraphics[scale=0.7]{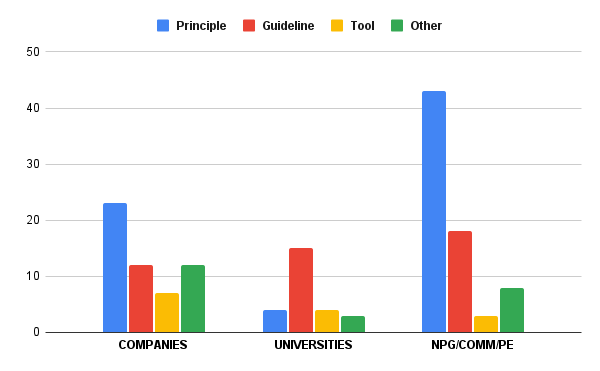}
\caption{Distribution of frameworks by category.}
\label{fig:research_results_typology_distribution}
\end{figure}

Furthermore, the frameworks have been classified into four categories according to their characteristics:
\begin{itemize}
    \item \textbf{Principle (P)}: if they highlight only abstract ethical principles or moral values;
    \item \textbf{Guideline (G)}: if they are concrete guidelines, quickly translatable into design constraints or choices;
    \item \textbf{Tool (T)}: if they are able to verify the compliance towards one or more principles and/or support practitioners in the implementation of principles or guidelines;
    \item \textbf{Other (O)}: if a resource cannot be classified into any of these categories - e.g. a list of possible attacks against an AI algorithm or a list of several questions to check in the design phase.
\end{itemize}


In Fig. \ref{fig:research_results_typology_distribution} the distribution of frameworks by their category and grouped by proposing institution is depicted\footnote{Differently from the previous graph, here we counted all frameworks, even if they are proposed by the same entity.}. 

It is noteworthy that COMPANIES and NPG/COMM/PE mostly proposed \textbf{Principles} (respectively 42.59\%, and 59.72\%); differently from this trend, UNIVERSITIES primarily proposed \textbf{Guidelines} (57.69\%). Furthermore, w.r.t. the number of resources proposed, the percentage of \textbf{Tools} proposed by COMPANIES is 12.96\% and 15.38\% for UNIVERSITIES, while this number is very low for NPG/COMM/PE (4.17\%).

\textbf{RQ2.} \emph{How much do these frameworks address the various RAI principles?}

In our rapid review, we are mainly interested in analyzing frameworks that give practical  support to all stakeholders involved in the development and deployment of AI applications. For this reason, to answer RQ2, we discarded, while computing our statistics, frameworks in the category \textit{Principle}, which cover only ethical values and that do not give any workable advice.
Furthermore, here we counted the frameworks and not the entities: if an entity proposed two (or more) frameworks, it was counted twice (or more).

Finally, before diving into the statistics, we point out that here we consider a principle as "covered" even if it is only "partially" covered, i.e. if not every aspect related to that principle was addressed.
For instance, we consider frameworks that deal with privacy but only in terms of how data is acquired from users and stored, without dealing with the new "privacy attacks" which can be conducted against an AI algorithm, like "model inversion" attack \cite{10.1145/2810103.2813677}.


\begin{figure}[ht]
\centering
\includegraphics[scale=0.7]{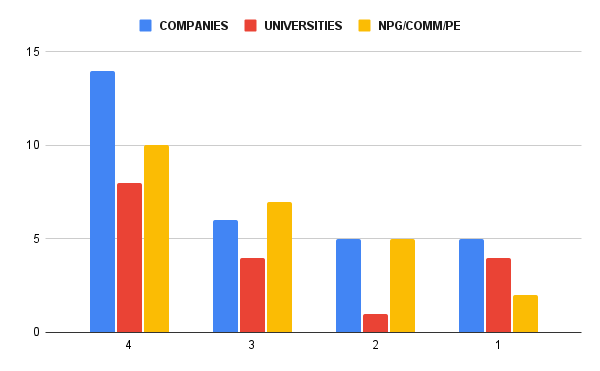}
\caption{Number of RAI principles addressed by the frameworks grouped by proposing entity type.}
\label{fig:research_results_num_RAI_principles_by_typology_distribution}
\end{figure}

As shown in Fig. \ref{fig:research_results_num_RAI_principles_by_typology_distribution}, in the majority of cases, the frameworks address all four RAI principles. Nevertheless, there are frameworks that cover only one (15.49\%) or two principles (15.49\%).
In particular, when the principles addressed are three, the most covered are \textit{Diversity \& Non-discrimination \& Fairness, Privacy \& Data Governance, and Transparency} (9.3\% for COMPANIES, 15.8\% for UNIVERSITIES and 14.1\% for NPG/COMM/PE).
While, for two principles, the most covered are \textit{Diversity \& Non-discrimination \& Fairness and Transparency} (9.3\% for COMPANIES and 16.9\% for NPG/COMM/PE). 
The interested reader can refer to the appendix for the complete results.

\textbf{RQ3.} \emph{Do these frameworks provide recommendations for each phase of the Software Development Life Cycle (SDLC)?}

\begin{figure}[ht]
\centering
\includegraphics[scale=0.7]{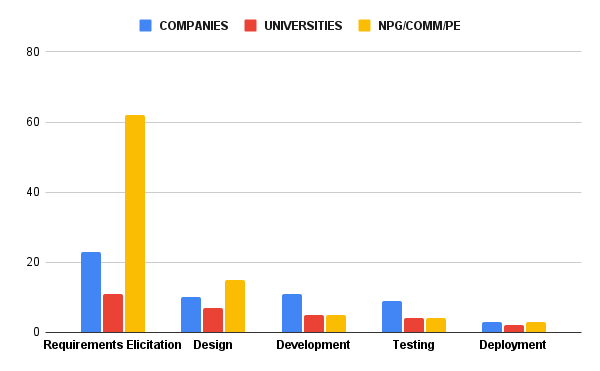}
\caption{Distribution by SDLC phase addressed.}
\label{fig:research_results_sdlc_distribution}
\end{figure}

To calculate the statistics to answer RQ3, RAI principles have been mapped to project requirements (see Section \ref{subsec:chosen_definition}) since they are actually high-level statements which put constraints on the project design. Moreover, we grouped resources proposed by each entity and mapped it w.r.t. the correspondent SDLC phase addressed. So, here, if an entity proposed two (or more) frameworks, it was counted once.

Fig. \ref{fig:research_results_sdlc_distribution} shows that there is a common pattern regardless of the type of the proposing entity: more than half of the frameworks provide support only for the \textbf{Requirements Elicitation} phase (96/174, 55.17\%), while subsequent phases are supported by very few frameworks. Anyway, COMPANIES showed more interest w.r.t. the other entities in also covering \textbf{Development} and \textbf{Testing} phases.
Finally, the \textbf{Deployment} phase is the least supported by the frameworks, regardless of the type of proposing entity.

\begin{figure}[ht]
\centering
\includegraphics[scale=0.7]{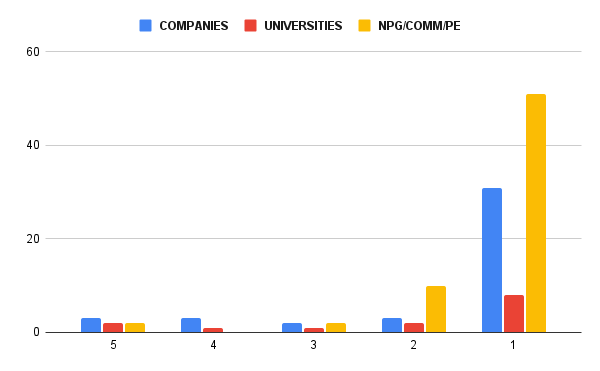}
\caption{Number of SDLC phases addressed by the frameworks grouped by proposing entity type.}
\label{fig:research_results_num_SDLC_phases_by_typology_distribution}
\end{figure}

In Fig. \ref{fig:research_results_num_SDLC_phases_by_typology_distribution} we investigate how much of the SDLC is covered (i.e. how many phases are addressed) by each framework.
Here, we counted the frameworks, not only the proponent entities. 
For frameworks proposed by COMPANIES sometimes we were not able to perform this mapping as no details on the internal core operations were provided.
This is due to the fact that they offered the algorithm audit service as a lucrative service, so they are "closed-source" frameworks. In such cases, we excluded the frameworks from the statistics elaboration.


What emerges is clear: most frameworks provide suggestions for just one of the five SDLC phases: \textbf{Requirements Elicitation}, while the second most supported phase \textbf{Design}.
Only 5.79\% of the frameworks supports all the SDLC phases. 

\textbf{RQ4.} \emph{Is there a supporting tool for each proposed framework?}

We highlight that in our analysis the presence of a supporting tool can be labeled with four values: "Yes", "No", "Partially" and "Not Mentioned".

\textbf{Partially} is used for:
\begin{itemize}
    \item Resources that cannot be properly considered tools; e.g. Excel sheets
    \item Tools that cover only a small part of the entire knowledge provided by the framework; e.g. a tool which covers only \textbf{Development} phase and neglects \textbf{Requirements} and \textbf{Design} phases
\end{itemize}
In the computation of values, we ignored the frameworks containing \textbf{Not mentioned} values for tools.

\begin{figure}[ht]
\centering
\includegraphics[scale=0.7]{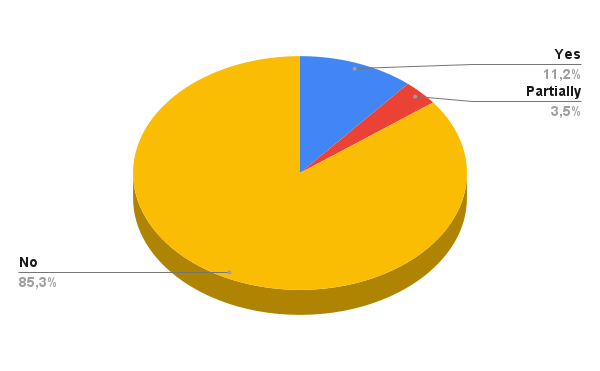}
\caption{Distribution by the existence of a supporting tool regardless of the proposing entity.}
\label{fig:research_results_tool_presence_distribution_no_entity_type}
\end{figure}

\begin{figure}[ht]
\centering
\includegraphics[scale=0.7]{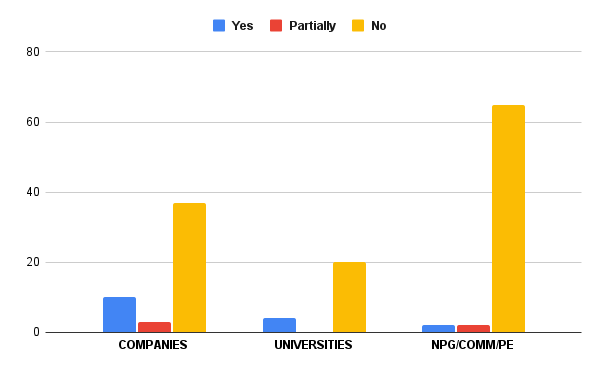}
\caption{Distribution by the existence of a supporting tool and proposing entity.}
\label{fig:research_results_tool_presence_distribution}
\end{figure}

Regarding the statistics, the results are quite clear: as shown in Fig. \ref{fig:research_results_tool_presence_distribution_no_entity_type} in more than the 80\% of the cases there is no tool included in the framework.
This is true regardless of the proposing entity as shown in Fig. \ref{fig:research_results_tool_presence_distribution}.

\begin{figure}[ht]
\centering
\includegraphics[scale=0.7]{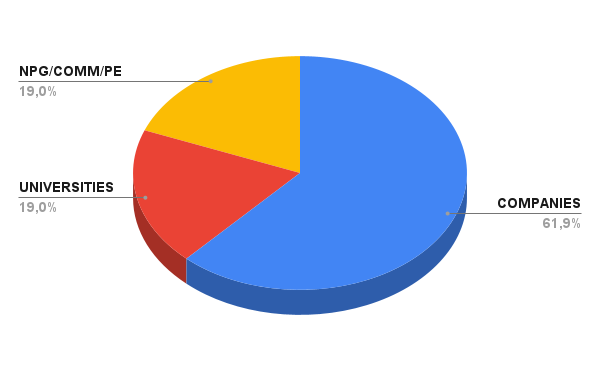}
\caption{Distribution by proposing entity in case a tool is provided.}
\label{fig:research_results_tool_presence_yes_distribution_by_entity_type}
\end{figure}

Moreover, Fig. \ref{fig:research_results_tool_presence_yes_distribution_by_entity_type} shows, when a tool is present, by which type of entity it is provided. Here the interesting result is that more than half of the tools are proposed by COMPANIES (61.9\%, 13/21).

\begin{figure}[ht]
\centering
\includegraphics[scale=0.7]{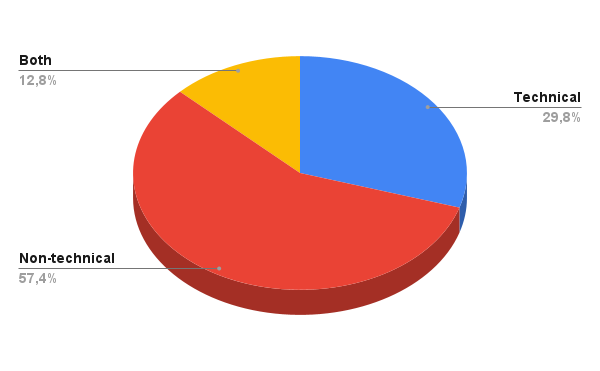}
\caption{Distribution of stakeholder's required background regardless of proposing entity in case a tool is provided.}
\label{fig:research_results_stakeholder_background_no_entity_type}
\end{figure}

Then, to go deeper into our investigation, we analyzed which kind of stakeholder's background is required to use these tools.
 We used the tag \textbf{Technical} to describe tools that should be used by stakeholders who work in the last three phases of the SDLC (e.g. developers, testers, IT technicians, etc); while we used the tag \textbf{Non-technical} for tools that should be used by stakeholders who work in the initial two phases of the SDLC (e.g. commercial agents, functional analysts, architecture designers, etc). Finally, if there are tools that offer resources for both kinds of stakeholders, we used the tag \textbf{Both}.
 Fig. \ref{fig:research_results_stakeholder_background_no_entity_type} shows that, regardless of the type of proposing entity, more than half of the tools can be used by \textbf{Non-technical} stakeholders (57.4\%, 27/47). Nevertheless, a significant amount of tools for supporting the \textbf{Technical} stakeholders is also provided (29.8\%, 14/47). The tools that can support both kinds of stakeholders, conversely, represent only a very low percentage (12.8\%, 6/47).

\begin{figure}[ht]
\centering
\includegraphics[scale=0.7]{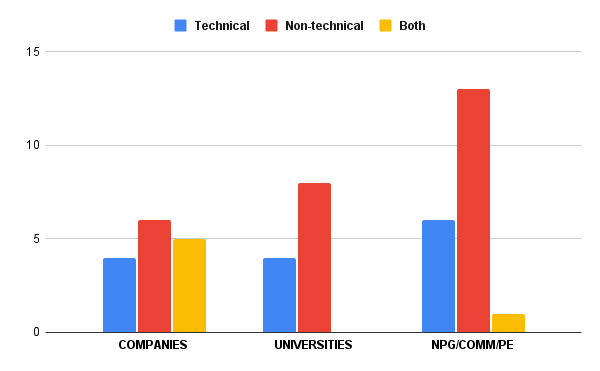}
\caption{Distribution of stakeholder's required background by proposing entity in case a tool is provided.}
\label{fig:research_results_stakeholder_background_by_entity_type}
\end{figure}

Furthermore, Fig. \ref{fig:research_results_stakeholder_background_by_entity_type} shows that a significant amount of tools proposed by COMPANIES can be used by both types of stakeholders (33.33\%) while UNIVERSITIES and NPG/COMM/PE proposed mainly tools that can be used by only one type of stakeholders.
\section{Discussion}
\label{sec:discussion}
\textbf{RQ1. What are the Responsible AI frameworks proposed in the literature?}
We have seen how the RAI frameworks found in our search are proposed by different types of entities.
 The difference in the percentages shown in Fig. \ref{fig:research_results_type_proposing_institution_distribution}, depending on the type of entity, may be due to various reasons, among others: the different amount of funds available for research (and implementation) and the Return-on-investment (ROI) of these kinds of initiatives, e.g. different between companies and no-profit organizations. Moreover, it is noteworthy that the group of NPG/COMM/PE is the largest and most heterogeneous group among the three, including as well scientific works conducted as academy-industry collaborations.

Regarding the way in which the frameworks are categorized, we can definitely say that there is a worrying lack of tools: most of the frameworks provide just \textbf{Principles} or \textbf{Guidelines}. We want to point this out since it is difficult to translate principles into concrete and measurable actions \cite{Mittelstadt2019} but a tool may help reduce this gap. Furthermore, without tools or well-known practices, it is not possible to measure the goodness and sufficiency of the adopted practices.

\textbf{RQ2. How much do these frameworks address the various RAI principles?}
In addressing this research question our aim was to discover how much "coverage" the frameworks give in terms of ethical principles handled. 
Indeed, sometimes practitioners are forced to use more than one framework at the same time in order to take into account the different dimensions of the RAI they want to implement and/or validate. 

The emerging trend seems positive: the majority of the frameworks address all the four most important RAI principles, even if sometimes in a "partial" way (the principle is not fully-covered, see Section \ref{sec:results}). 
Nevertheless, there are frameworks that neglect one, two, or even three principles. 
This could be due to multiple reasons, first of all, the fact that there is no global consensus about a clear definition of RAI and about which principles should be covered or implemented among the others.
Moreover, especially for companies, there could be reasons leading to ignoring part of the principles: among these, the \textit{Ethics Bluewashing} 
 \cite{Floridi2019five_risks}
 (the unethical behavior of making unsupported or false claims about, or taking insufficient action to support, the ethical ideals and advantages of digital processes, products, services, or other solutions in an effort to look more ethically conscientious than one actually is)
 and the \textit{Digital ethics shopping} (the unethical practice of selecting, modifying, or updating ethical guidelines, codes, frameworks, or other similar standards from a variety of offers in order to retrofit some pre-existing behaviors and thereby justify them a posteriori, rather than putting new behaviors into practice or improving them by comparing them to accepted ethical norms \cite{Floridi2019five_risks}).

\textbf{RQ3. Do these frameworks provide recommendations for each phase of the Software Development Life Cycle (SDLC)? }
The RR highlighted that most frameworks focus only on the initial phases of the SDLC, and in particular on \textit{Requirements elicitation}. This is a problem because all the statements and obligations provided as RAI principles must result in operations and practices actually implemented while developing an AI application, encompassing the entire SDLC. Indeed, for developers, testers, and system installers is really difficult to identify practices that can satisfy these abstract requirements without any support from an expert in the field or from a practical framework guiding them in all the development phases. What sounds alarming is a common trend across all the different types of entities, as the highest percentages of proposed frameworks fall in the \textit{Requirements Elicitation} phase while, from \textit{Design} onward, the numbers drop dramatically until the lows in the \textit{Deployment} phase.

\textbf{RQ4. Is there a supporting tool for each proposed framework?}
Here there is an emerging negative trend: in most cases, there is not a practical tool complementing the theoretical framework proposed and this is valid regardless of the type of entity releasing the tool.

 Moreover, most of the tools are Excel sheets, checklists, or questions to self-ask in the initial phases of the SDLC, so tools are mainly useful just for non-technical stakeholders. However, differently from the other entities, companies proposed a significant amount of tools useful for both kinds of stakeholders: technical and non-technical ones. This may be due to an interest in trying to sell complete services to other companies supporting every kind of stakeholder and, at the same time, covering every phase of the SDLC.

\textbf{Summary of findings.}
A global perspective on the results analyzed leads us to affirm that there is not a "catching-all" framework, simple and complete enough to provide different levels of knowledge for different kinds of stakeholders (technical and non-technical ones), which can simplify and speed up the adoption of RAI practices. This statement comes from the fact that there are no frameworks encompassing all RAI concepts logically and uniformly: recommendations are scattered across various pages or documents, and there is no one element that acts as a glue and makes the information consultation linear and fluid, at various levels of abstraction, for various types of stakeholders. To better explain this concept, we can take the work of Baldassarre et al. \cite{Baldassarre2019PrivacyOS} as an example: here, each element which composes the knowledge base is connected and this makes it possible to consult the knowledge base from various entry points, following different mental paths.
As a consequence, we advise the need for the creation of a uniform framework also for RAI: this way, the integration of RAI practices would be far easy in every kind of organization, public or private.
\section{Threats to validity}
\label{sec:threats}

In this section, we proceed to discuss the threats to the validity \cite{Wohlin2000ExperimentationIS} of our study. As exposed in \cite{10.1145/3210459.3210462}, Rapid Reviews usually present more threats to validity than other secondary studies due to their lightweight methodology.
\textit{Threats to internal validity} describe factors that could affect the results obtained from the study.\\
(i) Only two white-literature search engines were used, which may limit the number of primary studies found. To mitigate this, an in-depth analysis of grey-literature sources was performed.\\
(ii) To mitigate bias in the selection and extraction procedure, the process was conducted by two of the authors in a blind-view manner. Moreover, to impartially solve the conflicts, a third author intervened and took a decision. However, it is not possible to completely mitigate the bias threat.\\
(iii) With several iterations, including the refinement of keyword terms and search strategy, the authorship team attempted to capture as much relevant literature as possible. However, it is still possible that the authors’ chosen keywords and search strategy did not capture all relevant articles. For example, if a framework proposed by a company is not indexed on the used search engine and it is not simple to find a reference on the company's website, the framework may have been overlooked. However, given the number of results returned by the various grey-literature databases used, the authors feel confident that a representative amount of literature was captured through the search strategy described in Section \ref{sec:Study Design}.
\section{Conclusions}
\label{sec:conclusions}
In this work, we conduct a rapid review to provide an overview of the frameworks proposed in white and grey literature to help and speed up the adoption of Responsible Artificial Intelligence (RAI) practices. We identified four research questions to obtain specific answers w.r.t. our research goal, thus qualifying this survey's information.
To do a deeper investigation, we classified the entities providing each framework into three categories: COMPANIES, UNIVERSITIES, and NPG/COMM/PE. In this way, we were able to slice the results by the proposing entity.
This review focuses not only on scientific articles but includes also grey literature resources.

We can summarize the findings of this study as follows:\\
- \textbf{F1: Diversity of perspectives vs lack of standardization.} Frameworks of RAI are provided by multiple and heterogeneous entities, both public and private. This is, from one side, an added value as it contributes to the diversity of perspectives and to the so-called \textit{AI democratization}. 
Nevertheless, on the other side, this reveals an even greater lack of consensus and standardization about which are the best practices to follow to be compliant with the RAI ethical values.

- \textbf{F2: Theoretical vs practical support for AI practitioners.}
In our review, we also classified each framework according to the SDLC phases covered.
Our analysis highlights that very few frameworks encompass all the SDLC phases and provide practical support to practitioners who want to develop, test, and deploy RAI applications. There is a shortage of practical validation techniques for the theoretical principles, as well as implementation guidelines.
Moreover, there is a worrying deficit of tools supporting all the stakeholders in the implementation and auditing phase.

- \textbf{F3: No comprehensive framework.}
Even if a great number of the frameworks we analyzed cover all four selected principles, we observed that sometimes principles are only partially covered, reflecting the lack of standardization highlighted in F1.
In this rapid review what emerged is that current literature and industry lack complete, uniform, organized, and simple-to-use frameworks for RAI able to support the stakeholders across the entire Software Development Life Cycle (SDLC).
We can conclude that no comprehensive framework exists right now, whose knowledge can be navigated and exploited by different kinds of stakeholders (technical and
non-technical ones), which can simplify and speed up the
adoption of RAI practices. 

In conclusion, there is an undeniable agreement from different parts of the society about the need for ethical AI. However, we are still far from its realization, we lack consensus, standards, and tools: a large part of the road still needs to be covered before reaching this vision.
Therefore, further research must be conducted to develop such frameworks, which could speed up and give a practical boost to the adoption of RAI practices in real-world projects.

\bibliographystyle{unsrtnat}
\bibliography{references}  






\end{document}